\documentclass[11pt,a4paper]{article}
\usepackage[hyperref]{acl2019}
\usepackage{times}
\usepackage{latexsym}
\usepackage{graphicx}
\usepackage{amsmath}
\usepackage{multirow}
\usepackage{booktabs}
\usepackage{boldline}
\usepackage{float}

\usepackage{url}

\aclfinalcopy 
\def\aclpaperid{1234} 

\title{A Novel Bi-directional Interrelated Model for Joint Intent Detection and Slot Filling}
\author{
Haihong E\thanks{\indent Authors contributed
equally.},
Peiqing Niu\footnotemark[1],
Zhongfu Chen\footnotemark[1],
Meina Song\\
Beijing University of Posts and Telecommunications, Beijing, China\\
\texttt{\{ehaihong,niupeiqing,chenzhongfu,mnsong\}@bupt.edu.cn}\\
}
\begin{document}
\maketitle
\begin{abstract}
A spoken language understanding (SLU) system includes two main tasks, slot filling (SF) and intent detection (ID). The joint model for the two tasks is becoming a tendency in SLU. But the bi-directional interrelated connections between the intent and slots are not established in the existing joint models. In this paper, we propose a novel bi-directional interrelated model for joint intent detection and slot filling. We introduce an SF-ID network to establish direct connections for the two tasks to help them promote each other mutually. Besides, we design an entirely new iteration mechanism inside the SF-ID network to enhance the bi-directional interrelated connections. The experimental results show that the relative improvement in the sentence-level semantic frame accuracy of our model is 3.79\% and 5.42\% on ATIS and Snips datasets, respectively, compared to the state-of-the-art model.
\end{abstract}

\section{Introduction}
\begin{figure*}
	\centering
	\includegraphics[width=0.85\linewidth]{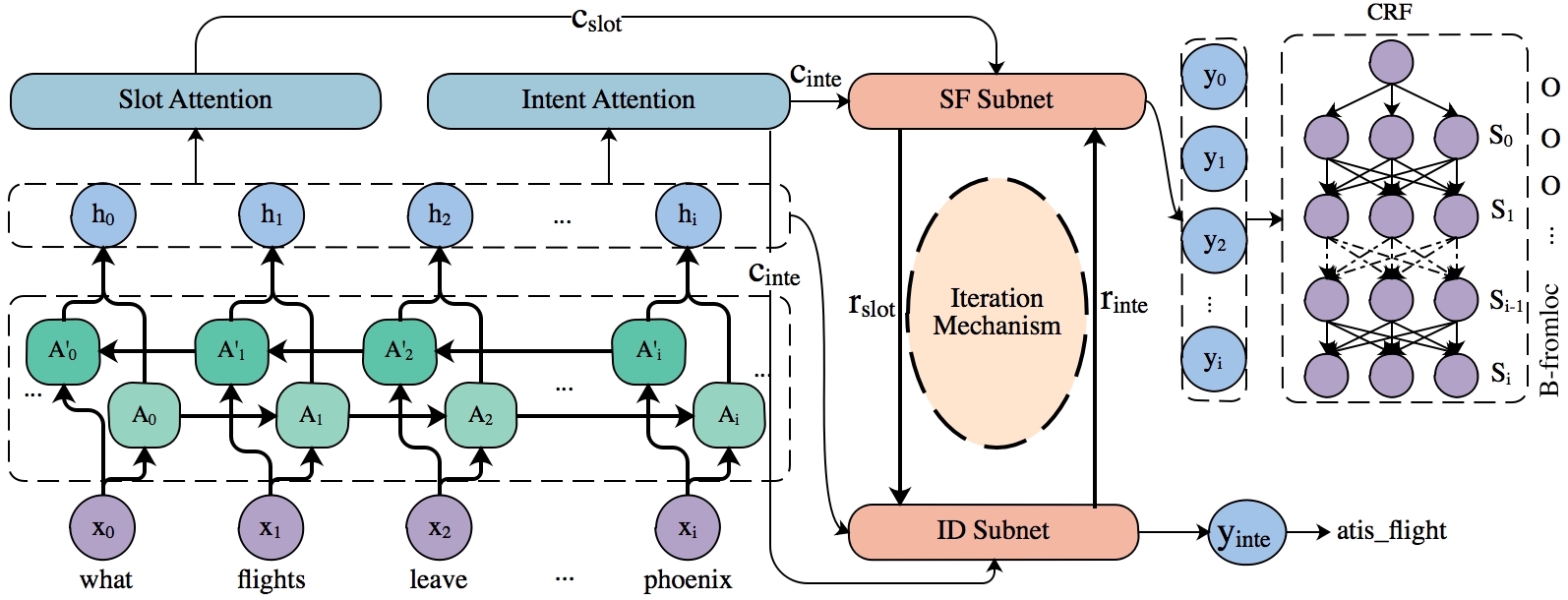}
	\caption{\label{fig:hist}The structure of the proposed model based on SF-ID network}
\end{figure*}
Spoken language understanding plays an important role in spoken dialogue system. SLU aims at extracting the semantics from user utterances. Concretely, it identifies the intent and captures semantic constituents. These two tasks are known as intent detection and slot filling \cite{tur2011spoken}, respectively. For instance, the sentence \textsl{`what flights leave from phoenix'} sampled from the ATIS corpus is shown in Table 1. It can be seen that each word in the sentence corresponds to one slot label, and a specific intent is assigned for the whole sentence.
\begin{table}[ht]
	\centering
	\small
	\begin{tabular}{|c|c|c|c|c|c|}
		\hline
		\textbf{Sentence}&what&flights&leave&from&phoenix\\
		\hline
		\textbf{Slots}&O&O&O&O&B-fromloc\\
		\hline
		\textbf{Intent}&\multicolumn{5}{|c|}{atis\_flight}\\
		\hline
	\end{tabular}
	\caption{\label{font-table}An example sentence from the ATIS corpus}
\end{table}

Traditional pipeline approaches manage the two mentioned tasks separately. Intent detection is seen as a semantic classification problem to predict the intent label. General approaches such as support vector machine (SVM) \cite{haffner2003optimizing} and recurrent neural network (RNN) \cite{lai2015recurrent} can be applied. Slot filling is regarded as a sequence labeling task. Popular approaches include conditional random field (CRF) \cite{raymond2007generative}, long short-term memory (LSTM) networks \cite{yao2014spoken}.

Considering the unsatisfactory performance of pipeline approaches caused by error propagation, the tendency is to develop a joint model \cite{chen2016syntax,zhang2016joint} for intent detection and slot filling tasks. \citeauthor{liu2016attention} (\citeyear{liu2016attention}) proposed an attention-based RNN model. However, it just applied a joint loss function to link the two tasks implicitly. \citeauthor{hakkani2016multi} (\citeyear{hakkani2016multi}) introduced a RNN-LSTM model where the explicit relationships between the slots and intent are not established. \citeauthor{goo2018slot} (\citeyear{goo2018slot}) proposed a slot-gated model which applies the intent information to slot filling task and achieved superior performance. But the slot information is not used in intent detection task. The bi-directional direct connections are still not established. In fact, the slots and intent are correlative, and the two tasks can mutually reinforce each other. This paper proposes an SF-ID network which consists of an SF subnet and an ID subnet. The SF subnet applies intent information to slot filling task while the ID subnet uses slot information in intent detection task. In this case, the bi-directional interrelated connections for the two tasks can be established. Our contributions are summarized as follows: 1) We propose an SF-ID network to establish the interrelated mechanism for slot filling and intent detection tasks. Specially, a novel ID subnet is proposed to apply the slot information to intent detection task. 2) We establish a novel iteration mechanism inside the SF-ID network in order to enhance the connections between the intent and slots. 3) The experiments on two benchmark datasets show the effectiveness and superiority of the proposed model. 
\section{Proposed Approaches}
This section first introduces how we acquire the integration of context of slots and intent by attention mechanism. And then it presents an SF-ID network which establishes the direct connections between intent and slots. The model architecture based on bi-directional LSTM (BLSTM) is shown in Figure 2.$\footnote{The code is available at \url{https://github.com/ZephyrChenzf/SF-ID-Network-For-NLU}}$
\subsection{Integration of Context}
In SLU, word tags are determined not only by the corresponding terms, but also the context \cite{chen2016end}. The intent label is also relevant with every element in the utterance. To capture such dependencies, attention mechanism is introduced.\\
\textbf{Slot filling:} The $i^{th}$ slot context vector $c^i_{slot}$ is computed as the weighted sum of BLSTM's hidden states $(h_1,...,h_t)$:
\begin{equation} c^i_{slot}=\sum_{j=1}^T{\alpha^S_{i,j}h_j} \end{equation}
where the attention weight $\alpha$ is acquired the same way as in \cite{liu2016attention}.\\
\textbf{Intent detection:} The intent context vector $c_{inte}$ is calculated as the same way as $c_{slot}$, in particular, it just generates one intent label for the whole sentence.
\subsection{SF-ID Network}
The SF-ID network consists of an SF subnet and an ID subnet. The order of the SF and ID subnets can be customized. Depending on the order of the two subnets, the model have two modes: SF-First and ID-First. The former subnet can produce active effects to the latter one by a medium vector.
\subsubsection{SF-First Mode}
In the SF-First mode, the SF subnet is executed first. We apply the intent context vector $c_{inte}$ and slot context vector $c_{slot}$ in the SF subnet and generate the slot reinforce vector $r_{slot}$. Then, the newly-formed vector $r_{slot}$ is fed to the ID subnet to bring the slot information.\\
\textbf{SF subnet}: The SF subnet applies the intent and slot information (i.e. $c_{inte}$ and $c_{slot}$) in the calculation of a correlation factor $f$ which can indicate the relationship of the intent and slots. This correlation factor $f$ is defined by:
\begin{equation} f=\sum{V*tanh(c_{slot}^i+W*c_{inte})} \end{equation}
In addition, we introduce a slot reinforce vector $r_{slot}$ defined by (3), and it is fed to the ID subnet to bring slot information.
\begin{equation} r_{slot}^i=f \cdot c_{slot}^i\end{equation}
\textbf{ID subnet}: We introduce a novel ID subnet which applies the slot information to the intent detection task. We believe that the slots represent the word-level information while the intent stands for the sentence-level. The hybrid information can benefit the intent detection task. The slot reinforce vector $r_{slot}$ is fed to the ID subnet to generate the reinforce vector $r$, which is defined by:
\begin{equation} r=\sum_{i=1}^{T}{\alpha_i\cdot r_{slot}^i} \end{equation}
where the weight $\alpha_i$ of $r_{slot}^i$ is computed as:
\begin{gather} 
\alpha_i=\frac{exp(e_{i,i})}{\sum^T_{j=1}exp(e_{i,j})} \\
e_{i,j}=W*tanh(V_1*r_{slot}^i+V_2*h_j+b) 
\end{gather}
We also introduce an intent reinforce vector $r_{inte}$ which is computed as the sum of the reinforce vector $r$ and intent context vector $r_{inte}$. 
\begin{equation} r_{inte}= r+c_{inte} \end{equation}
\textbf{Iteration Mechanism}: The intent reinforce vector $r_{inte}$ can
\begin{figure}
 \centering
 \includegraphics[width=0.8\linewidth]{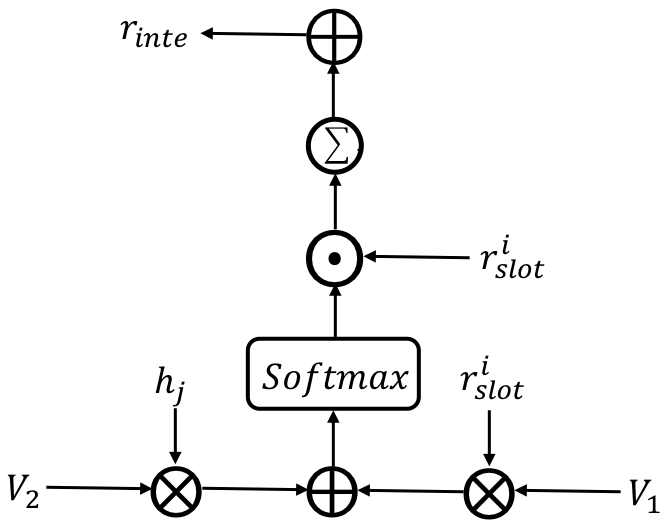}
 \caption{\label{fig:hist}Illustration of the ID subnet}
\end{figure}
also be fed into the SF subnet. In fact, this intent reinforce vector $r_{inte}$ can improve the effect of relation factor $f$ because it contains the hybrid information of intent and slots, and (2) can be replaced by:
\begin{equation} f=\sum{V*tanh(c_{slot}^i+W*r_{inte})} \end{equation}
With the change in the relation factor $f$, a new slot reinforce vector $r_{slot}$ is acquired. Thus, the ID subnet can takes a new $r_{slot}$ and exports a new $r_{inte}$. In this case, both SF subnet and ID subnet are updated, one iteration is completed.

In theory, the interaction between the SF subnet and ID subnet can repeat endlessly, which is denoted as the iteration mechanism in our model. The intent and slot reinforce vectors act as the links between the SF subnet and the ID subnet and their values continuously change during the iteration process.

After the iteration mechanism, the $r_{inte}$ and $r_{slot}$ participate in the final prediction of intent and slots, respectively. For the intent detection task, the intent reinforce vector $r_{inte}$ and the last hidden state $h_T$ of BLSTM are utilized in the final intent prediction:
\begin{equation}
y_{inte}=softmax(W_{inte}^{hy}concat(h_T,r_{inte}))
\end{equation}
For the slot filling task, the hidden state $h_i$ combined with its corresponding slot reinforce vector $r_{slot}^i$ are used in the $i^{th}$ slot label prediction. The final expression without CRF layer is:
\begin{equation}
y_{slot}^i=softmax(W_{slot}^{hy}concat(h_i,r_{slot}^i))
\end{equation}
\subsubsection{ID-First Mode}
In the ID-First mode, the ID subnet is performed before the SF subnet. In this case, there are some differences in the calculation of ID subnet in the first iteration.\\
\textbf{ID subnet}: Unlike the Slot-First mode, the reinforce vector $r$ is acquired by the hidden states and the context vectors of BLSTM. Thus, (4) (5) (6) can be replaced by:
\begin{gather} 
r=\sum_{i=1}^{T}{\alpha_i\cdot h_i}\\
\alpha_i=\frac{exp(e_{i,i})}{\sum^T_{j=1}exp(e_{i,j})}\\
e_{i,j}=W*\sigma(V_1*h_i+V_2*c_{slot}^{j}+b) 
\end{gather}
The intent reinforce vector $r_{inte}$ is still defined by (7), and it is fed to the SF subnet.\\
\textbf{SF subnet}: The intent reinforce vector $r_{inte}$ is fed to the SF subnet and the relation factor $f$ is calculated the same way as (8).
Other algorithm details are the same as in SF-First mode.\\
\textbf{Iteration Mechanism}: Iteration mechanism in ID-First mode is almost the same as that in SF-First mode except for the order of the two subnets. 
\subsection{CRF layer}
Slot filling is essentially a sequence labeling problem. For the sequence labeling task, it is beneficial to consider the correlations between the labels in neighborhoods. Therefore, we add the CRF layer above the SF subnet outputs to jointly decode the best chain of labels of the utterance. 
\begin{table*}[ht]
		\centering
		\small
		\begin{tabular}{c|r|ccc|ccc}
			\hlineB{2}
			\multicolumn{2}{c|}{ \multirow{2}*{\textbf{Model}}}& \multicolumn{3}{c|}{\textbf{ATIS Dataset}} &\multicolumn{3}{c}{\textbf{Snips Dataset}}\\
			\cline{3-8}
			\multicolumn{2}{c|}{}&\textbf{Slot (F1)}&\textbf{Intent (Acc)}&\textbf{Sen. (Acc)}&\textbf{Slot (F1)}&\textbf{Intent (Acc)}&\textbf{Sen. (Acc)}\\
			\hline
			\multicolumn{2}{r|}{Joint Seq \cite{hakkani2016multi}}&94.30&92.60&80.70&87.30&96.90&73.20\\
			\multicolumn{2}{r|}{Atten.-Based \cite{liu2016attention}}&94.20&91.10&78.90&87.80&96.70&74.10\\
			\multicolumn{2}{r|}{Sloted-Gated \cite{goo2018slot}}&95.42&95.41&83.73&89.27&96.86&76.43\\
			\hline
			\hline
			\multicolumn{1}{c|}{ \multirow{4}*{}}&SF-First (with CRF)&95.75&\textbf{97.76}&86.79&91.43&\textbf{97.43}&\textbf{80.57}\\
			\multicolumn{1}{c|}{SF-ID}&SF-First (without CRF)&95.55&97.40&85.95&90.34&97.34&78.43\\
			\multicolumn{1}{c|}{Network}&ID-First (with CRF)&\textbf{95.80}&97.09&\textbf{86.90}&\textbf{92.23}&97.29&80.43\\
			\multicolumn{1}{c|}{}&ID-First (without CRF)&95.58&96.58&86.00&90.46&97.00&78.37\\
			\hlineB{2}
		\end{tabular}
		\caption{\label{font-table}Performance comparison on ATIS and Snips datasets. The improved cases are written in bold.  }
	\end{table*}
\section{Experiment}
\begin{table}[ht]
	\centering
	\small
	\begin{tabular}{ccccc}
		\toprule
		\multicolumn{1}{c}{\multirow{2}*{Model}}&\multicolumn{2}{c}{ATIS}&\multicolumn{2}{c}{Snips}\\
	    \cmidrule{2-5}
		\multicolumn{1}{c}{}&Slot&Intent&Slot&Intent\\
		\midrule
		Without SF-ID&95.05&95.34&88.9&96.23\\
		ID subnet Only&95.43&95.74&89.57&97.42\\
		SF subnet Only&95.14&95.75&90.7&96.71\\
		\midrule
		SF-ID (no interaction)&95.56&95.75&90.97&97.01\\
		SF-ID (SF-First)&95.75&\textbf{97.76}&91.43&\textbf{97.43}\\
		SF-ID (ID-First)&\textbf{95.80}&97.09&\textbf{92.23}&97.29\\
		\bottomrule
	\end{tabular}
	\caption{\label{font-table}Analysis of seperate subnets and their interaction effects}
\end{table}
\textbf{Dataset}: We conducted experiments using two public datasets,
the widely-used ATIS dataset \cite{hemphill1990atis} and custom-intent-engine dataset called the Snips \cite{coucke2018snips}, which is collected by Snips personal voice assistant. Compared with the ATIS dataset, the Snips dataset is more complex due to its large vocabulary and cross-domain intents.\\
\textbf{Evaluation Metrics}: We use three evaluation metrics in the experiments. For the slot filling task, the F1-score is applied. For the intent detection task, the accuracy is utilized. Besides, the sentence-level semantic frame accuracy (sentence accuracy) is used to indicate the general performance of both tasks, which refers to proportion of the sentence whose slots and intent are both correctly-predicted in the whole corpus.\\
\textbf{Training Details}: In our experiments, the layer size for the BLSTM networks is set to 64. During training, the adam optimization \cite{kingma2014adam} is applied. Besides, the learning rate is updated by $\eta_t=\eta_0/(1+pt)$ with a decay rate of $p=0.05$ and an initial learning rate of $\eta_0=0.01$, and $t$ denotes the number of completed steps.\\
\textbf{Model Performance}: The performance of the models are given in Table 2, wherein it can be seen that our model outperforms the baselines in all three aspects: slot filling (F1), intent detection (Acc) and sentence accuracy (Acc). Specially, on the sentence-level semantic frame results, the relative improvement is around 3.79\% and 5.42\% for ATIS and Snips respectively, indicating that SF-ID network can benefit the SLU performance significantly by introducing the bi-directional interrelated mechanism between the slots and intent.\\
\textbf{Analysis of Seperate Subnets}: We analyze the effect of seperate subnets, and the obtained results are given in Table 3. The experiments are conducted when the CRF layer is added. As we can see, both models including only the SF subnet or the ID subnet have acheived better results than the BLSTM model. Therefore, we believe that both SF subnet and ID subnet have significance in performance improvement.

Beside, we also analyse the condition with independent SF and ID subnet, in other words, when there is no interaction in SF and ID subnet. We can see it also obtains good results. However, the SF-ID network which allows the two subnets interact with each other achieve better results. This is because the bi-directional interrelated mechanism help the two subnets promote each other mutually, which improves the performance in both tasks.\\
\textbf{Analysis of Model Mode}: In Table 2, it can be seen that the ID-First mode achieves better performance in the slot filling task. This is because the ID-First mode treats the slot filling task as a more important task, because the SF subnet can utilize the intent information output from the ID subnet. Similarly, the SF-First mode performs better in the intent detection task. In general, the difference between the two modes is minor.\\
\textbf{Iteration Mechanism}: The effect of iteration mechanism is shown in Figure 3. 
\begin{figure}
	\centering
	\includegraphics[width=0.75\linewidth]{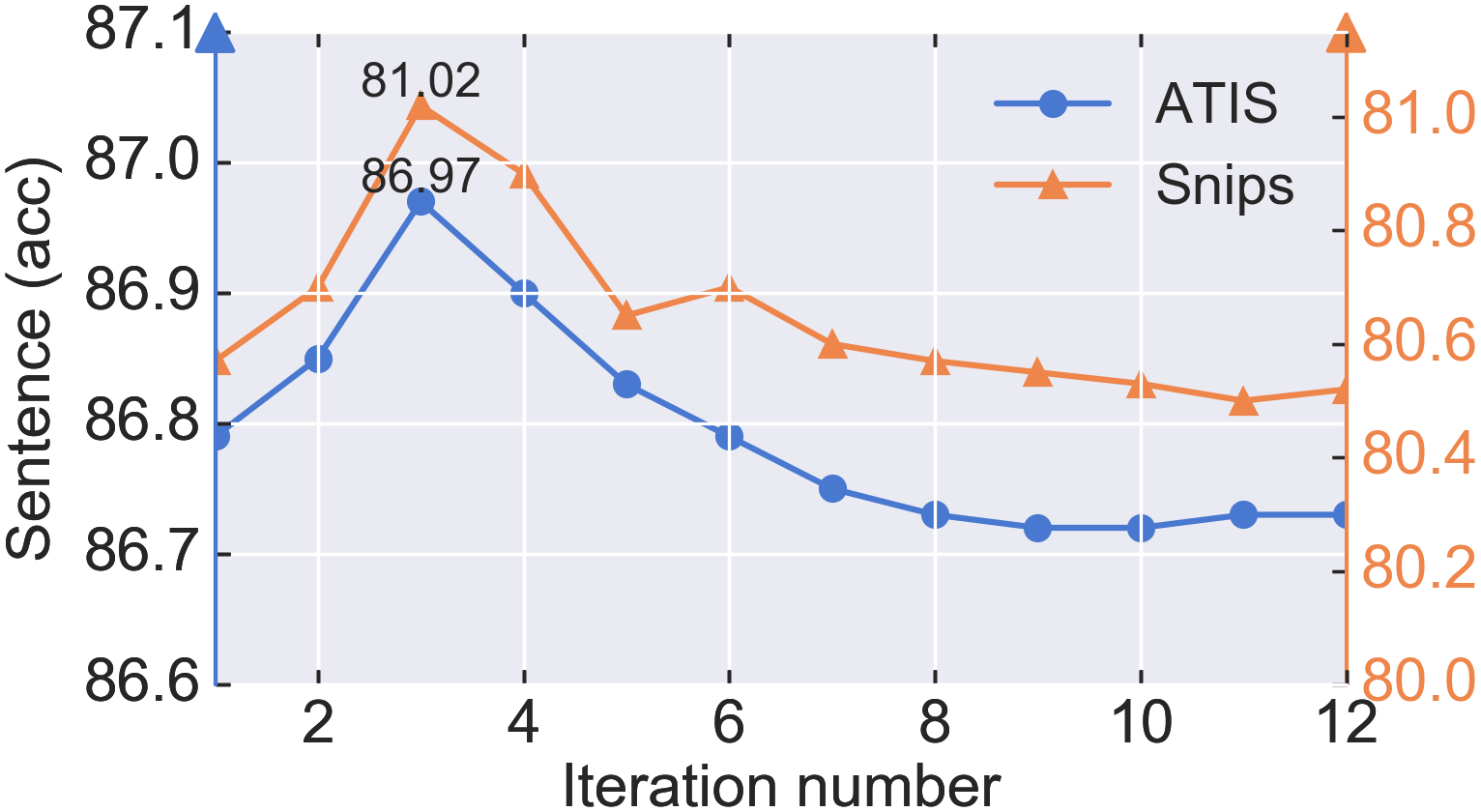}
	\caption{\label{fig:hist}Effect of iteration number on the model performance in SF-First mode}
\end{figure}
The experiments are conducted in SF-First mode. Sentence accuracy is applied as the performance measure because it can reflect the overall model performance. It increases gradually and reaches the maximum value when the iteration number is three on both ATIS and Snips dataset, indicating the effectiveness of iteration mechanism. It may credit to the iteration mechanism which can enhance the connections between intent and slots. After that, the sentence accuracy gradually gets stabilized with minor drop. On balance, the iteration mechanism with proper iteration number can benefit the SLU performance.\\
\textbf{CRF Layer}: From Table 2 it can be seen that the CRF layer has a positive effect on the general model performance. This is because the CRF layer can obtain the maximum possible label sequence on the sentence level. However, CRF layer mainly focuses on sequence labeling problems. So the improvement of the slot filling task obviously exceeds that of the intent detection task. In general, the performance is improved by the CRF layer.
\section{Conclusion}
In this paper, we propose a novel SF-ID network which provides a bi-directional interrelated mechanism for intent detection and slot filling tasks. And an iteration mechanism is proposed to enhance the interrelated connections between the intent and slots. The bi-directional interrelated model helps the two tasks promote each other mutually. Our model outperforms the baselines on two public datasets greatly. This bi-directional interrelated mechanism between slots and intent provides guidance for the future SLU work. 
\section*{Acknowledgments}
The authors would like to thank the reviewers for their valuable comments. This work was supported in part by the National Key R\&D Program of China under Grant SQ2018YFB140079 and 2018YFB1403003.
\bibliographystyle{acl_natbib}
\bibliography{acl2019}
\end{document}